\documentclass{article}

\usepackage[preprint]{neurips_2025}

\usepackage[utf8]{inputenc} 
\usepackage[T1]{fontenc}    
\usepackage{hyperref}       
\usepackage{url}            
\usepackage{booktabs}       
\usepackage{amsfonts}       
\usepackage{nicefrac}       
\usepackage{microtype}      
\usepackage{xcolor}         

\usepackage{amsmath}
\usepackage{graphicx}
\def\thickhline{\noalign{\hrule height1pt}}
\usepackage{array}
\usepackage{multirow}
\definecolor{linkblue}{HTML}{0000EE}
\newcommand{\brandc}[1]{\textcolor[HTML]{1447e6}{#1}}
\newcommand{\sm}[1]{{\scriptsize #1}}

\title{Improving Large Language Models with \\Concept-Aware Fine-Tuning}

\author{%
  Michael K. Chen \\
  Nanyang Technological University\\
  Singapore \\
  \texttt{michaelchenkj@gmail.com} \\
  \And
  Xikun Zhang \\
  Nanyang Technological University \\
  Singapore \\
  \texttt{xikun.zhang@ntu.edu.sg} \\
  \AND
  Jiaxing Huang \\
  Nanyang Technological University \\
  Singapore \\
  \texttt{jiaxing.huang@ntu.edu.sg} \\
  \And
  Dacheng Tao \\
  Nanyang Technological University \\
  Singapore \\
  \texttt{dacheng.tao@ntu.edu.sg} \\
}

\begin{document}

\maketitle

\begin{abstract}
  Large language models (LLMs) have become the cornerstone of modern AI. However, the existing paradigm of next-token prediction fundamentally limits their ability to form coherent, high-level concepts, making it a critical barrier to human-like understanding and reasoning. Take the phrase "ribonucleic acid" as an example: an LLM will first decompose it into tokens, i.e., artificial text fragments ("rib" → "on" → …), then learn each token sequentially, rather than grasping the phrase as a unified, coherent semantic entity. This fragmented representation hinders deeper conceptual understanding and, ultimately, the development of truly intelligent systems. In response, we introduce \brandc{\textbf{Concept-Aware Fine-Tuning (CAFT)}}, a novel multi-token training method that redefines how LLMs are fine-tuned. By enabling the learning of sequences that span multiple tokens, this method fosters stronger concept-aware learning. Our experiments demonstrate significant improvements compared to conventional next-token fine-tuning methods across diverse tasks, including traditional applications like text summarization and domain-specific ones like de novo protein design. Multi-token prediction was previously only possible in the prohibitively expensive pretraining phase; CAFT, to our knowledge, is the first to bring the multi-token setting to the post-training phase, thus effectively democratizing its benefits for the broader community of practitioners and researchers. Finally, the unexpected effectiveness of our proposed method suggests wider implications for the machine learning research community. All code and data are available at \href{https://github.com/michaelchen-lab/caft-llm}{\color{linkblue}{https://github.com/michaelchen-lab/caft-llm}}


\end{abstract}

\section{Introduction}


Large language models (LLMs) have advanced tremendously in recent years. They owe much of their success to the effectiveness of the LLM development pipeline \citep{li2024pre}, which can be described as such: first, in the pre-training phase, models are trained on a large-scale unsupervised text corpus in order to teach general knowledge and language understanding. Next, in the post-training phase, models are fine-tuned on downstream supervised datasets to respond to diverse tasks in specific formats and to prevent dangerous model behaviors through a myriad of techniques. This is done via a variety of techniques, including instruction tuning, reinforcement learning from human feedback (RLHF), and more \citep{lambert2024t}. This training paradigm has fueled the growth of language models in both AI research and commercial adoption.

\begin{figure}[ht!]
	\centering
	\includegraphics[width=1.0\textwidth]{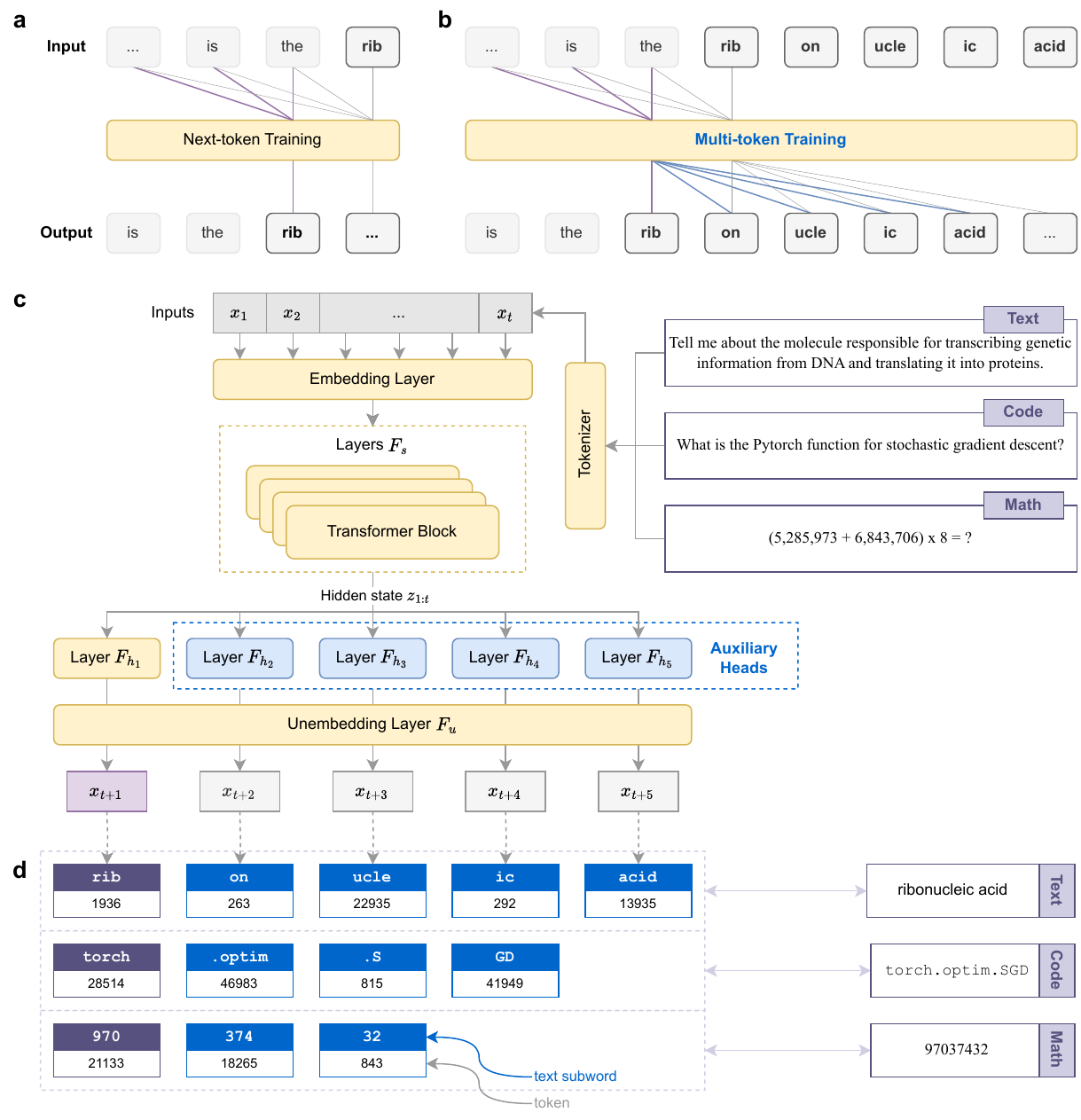}
	\caption{\textbf{(a, b) Next-token vs. multi-token training.} Language models are typically trained using the next-token objective (left), where each token is an artificial text fragment. At every forward pass, models are tasked to predict the next immediate token. However, in the multi-token setting (right), models are tasked to predict the next $n$ tokens in parallel in each forward pass, thus facilitating conceptual understanding across tokens. In this illustration, the relevant concept is ``ribonucleic acid", which is segmented into 5 tokens by the Llama 3 tokenizer. \textbf{(c) Concept-aware fine-tuning (CAFT) architecture.} Next-token models are modified into multi-token ones by training task-agnostic auxiliary heads $F_{h_k}$ (blue) that predict the next $k$-th token (Section \ref{subsec:method_aux_heads}) for $n=5$. During multi-token fine-tuning, the auxiliary heads augment the cross-entropy loss to enable the multi-token training objective (Section \ref{subsec:method_ft}). Finally, only the original model layers (in yellow) are used for inference. \textbf{(d) Examples of multi-token concepts.} Our proposed method increases coherence across domains and modalities, including text, code, and mathematical expressions, given the prevalence of multi-token concepts.}
	\label{fig:main}
\end{figure}

Importantly, this training paradigm conforms to a seemingly unassailable training objective: next-token prediction. A vocabulary of tokens, or text fragments, is first created using tokenization algorithms, most commonly byte-pair encoding (BPE) \citep{sennrich2015neural}, which forms word/subword tokens based on their frequency in the training corpus. After tokenizing the texts using this vocabulary, the tokens are fed into the model to predict the next token autoregressively. For example, as shown in Figure \ref{fig:main}(a,b), if a Llama 3 model \citep{grattafiori2024llama} is tasked to predict \textit{ribonucleic acid} as part of a given question, the phrase is first deconstructed, i.e., tokenized, into \textit{rib}, \textit{on}, \textit{ucle}, \textit{ic}, and \textit{acid}. Then, the model is trained to predict a single token in each forward pass sequentially, starting from \textit{rib}.

However, this training objective is suboptimal: tokens are artificial text fragments that do not represent coherent concepts or entities. At each forward pass, models have no access to the succeeding tokens. For example, when predicting \textit{rib} as part of \textit{ribonucleic acid}, the latter portion \textit{-onucleic acid} is hidden. Intuitively, learning a single token that is part of a larger concept, in isolation, fails to capture the actual underlying information at hand. The growing literature on the effects of tokenization on language model performance supports this assertion. Tokenizers with better compression, i.e., the ability to discretize text into longer words and subwords, lead to better model performances than those with worse compression \citep{goldman2024unpacking}. Additionally, the specific implementation of tokenization affects how numbers and mathematical expressions are segmented, ultimately having undue influence on arithmetic ability \citep{singh2024tokenization}. Training models to predict the next immediate token handicaps their learning process.

Instead, models should be trained to predict \textit{concepts}, often spanning multiple tokens, as shown in Figure \ref{fig:main}d. Along these lines, several methods leveraging multi-token prediction have been proposed \citep{gloeckle2024better, liu2024deepseek}. Specifically, at each position in the training corpus, models are trained to predict the following $n$ tokens using $n$ output heads. However, these methods are restricted to the \textit{pretraining} phase, which results in prohibitive costs and diminished effectiveness. First, the pretraining phase is inherently \textit{orders of magnitudes} more computationally expensive than post-training, making existing multi-token methods unfeasible for all but a select group of well-resourced labs. Second, the pretraining phase teaches models general knowledge and language modeling skills, while the post-training phase teaches specific, relevant skills. Thus, existing methods do not adequately learn domain-specific, multi-token concepts: they exhibit only incremental gains compared to their next-token counterparts on downstream tasks.

Naturally, one would expect multi-token prediction to be applied to fine-tuning instead. However, to the best of our knowledge, current research in this direction has been unsuccessful, finding that fine-tuning with multi-token prediction leads to similar or worse performance \citep{gloeckle2024better, cai2024medusa}. Incorporating the multi-token setting into the post-training phase is extremely challenging because the multi-token setting represents a dramatic distribution shift. Given that post-training is much shorter than pretraining, models fail to adapt, leading to degradation.

In response, we introduce \textbf{Concept-Aware Fine-Tuning (CAFT)}, a novel multi-token fine-tuning method for next-token models. First, auxiliary heads that predict token positions beyond the next immediate token are trained using an instruction-tuning mixture, where the ground truth responses are self-distilled from the model itself. We provide trained task-agnostic auxiliary heads for a range of popular open-source models, allowing practitioners to focus on their task-specific MTP fine-tuning, as illustrated in Figure \ref{fig:main}c. On top of full or Low-Rank Adaptation (LoRA) fine-tuning on the base model, the auxiliary heads and multi-token loss function are added.

We empirically demonstrate CAFT's effectiveness and applicability to diverse domains, including traditional ones like text summarization and domain-specific ones like de novo protein design. It achieves superior performance to its next-token full and LoRA fine-tuning counterparts. The magnitudes of gains are similar or better than existing MTP pretraining methods despite using only a fraction of the computational cost. Additionally, we find that CAFT LoRA often outperforms next-token full fine-tuning, suggesting that models learn more effectively in a multi-token setting. In settings where multi-token prediction is highly advantageous, a multi-fold increase in model performance can be observed.

Importantly, CAFT presents significant implications for the scientific community. First, by introducing multi-token prediction into the post-training phase, our method democratizes the benefits of MTP to the broader community of practitioners and researchers. This builds the foundation for future works in this nascent frontier. Second, LLMs' ability to plan beyond the next token is still hotly debated \citep{lindsey2025biology}. The unreasonable effectiveness of CAFT demonstrates that the models do not adequately learn and plan ahead of the next immediate token; an explicit multi-token objective is more effective. Our empirical evidence serves as a crucial step towards understanding the internal mechanisms of language models.

\section{Concept-Aware Fine-Tuning (CAFT)}

Auxiliary heads are first trained in order to facilitate multi-token fine-tuning, which we describe in Section \ref{subsec:method_aux_heads}. This only needs to be trained once for a given model and can be provided by a third-party, so practitioners need only focus on the next step: multi-token fine-tuning on their specific task, which is described in Section \ref{subsec:method_ft}. To better illustrate the multi-token setting, we first briefly describe the canonical next-token training method. 

\begin{figure}
	\centering
	\includegraphics[width=1.0\textwidth]{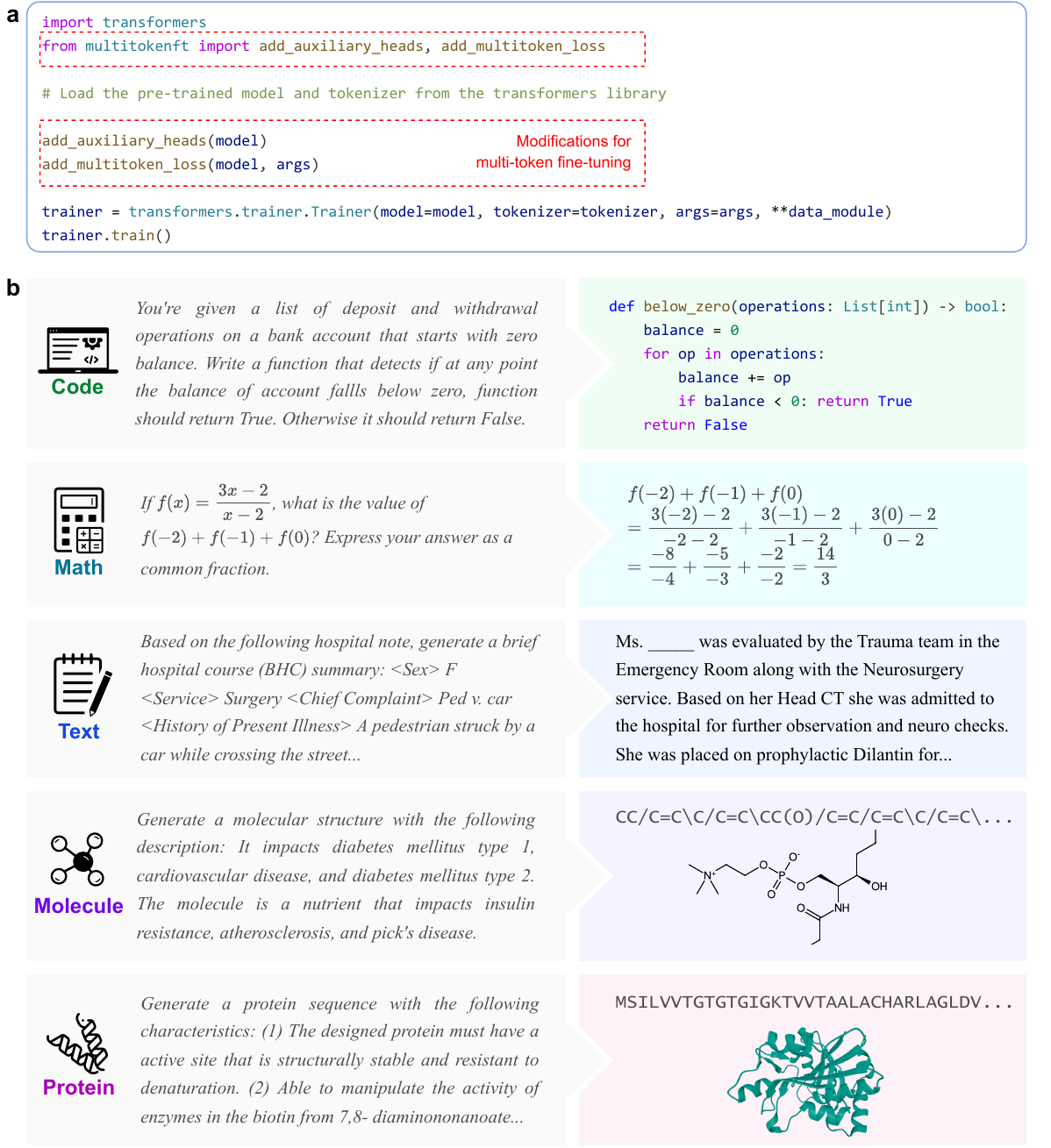}
	\caption{\textbf{(a) Sample code implementation.} Building on top of the industry-standard Transformers library \citep{wolf2020transformers}, researchers and practitioners can incorporate CAFT into their existing Transformers training pipelines with just a few lines of code using our open-source library \brandc{\texttt{caft}}. \textbf{(b) Downstream Tasks.} These tasks empirically underscore the effectiveness and broad applicability of CAFT. The examples are adapted from the HumanEval, MATH-500, MIMIC-IV-BHC, L+M-24, and Mol-Instructions datasets respectively, which are used for the evaluations in Section \ref{subsec:tasks_results}.}
	\label{fig:use_cases}
\end{figure}

\subsection{Background on Next-token Prediction}
\label{subsec:method_bg}

Conventionally, language models are trained autoregressively on a large text corpus using a next-token prediction task, as illustrated in Figure \ref{fig:main}a. Given the inputs $x_1$,...$x_t$, models are tasked to predict $x_{t+1}$ with the objective of minimizing the following cross-entropy loss:
\begin{equation}
\mathcal{L}_1 = \log p_{t} (y_{t+1})
\end{equation}

where $y_{t+1}$ is the ground truth token at position $t+1$. This core objective dominates both pretraining and fine-tuning (which is applied in the post-training phase). In this work, we reshape this ubiquitous training objective to predict the next $n$ tokens instead, as shown in Figure \ref{fig:main}b.

\textbf{The challenge of multi-token fine-tuning.} Incorporating the multi-token setting into the post-training phase is not a new idea; however, existing attempts have been unsuccessful. For example, \cite{gloeckle2024better} finds that next-token prediction performs better during fine-tuning—even for models pretrained under the multi-token setting! Naively repurposing multi-token pretraining methods to fine-tune next-token models creates several problems: First, introducing the multi-token objective represents a dramatic distribution shift, which models often do not recover from, leading to worse performances than the base models. Second, given the naturally higher loss of auxiliary heads (due to having further token position), models tend to optimize for the auxiliary loss at the expense of $\mathcal{L}_1$, which is ultimately what matters at inference time. Lastly, the post-training phase is dramatically shorter than pretraining; using existing methods, models do not have sufficient compute to take advantage of the multi-token setting during fine-tuning. 

Our proposed method, CAFT, introduces a series of novel techniques and applications designed to address these previously unsolved challenges, making it the first to enable multi-token fine-tuning.

\subsection{Setting the stage: Training auxiliary heads}
\label{subsec:method_aux_heads}

Before CAFT can be applied to next-token models, they must be adapted to predict $n$ future tokens at once. As such, auxiliary heads that predict the $k$-th token for $k=1,2,..,n$ are trained. Importantly, these heads are task-agnostic and can be used on a variety of downstream fine-tuning tasks.

$n-1$ auxiliary heads are added to predict the next $n$ tokens. The architecture consists of (i) an independent hidden layer $F_{h_k}$, whose weights are initialized identically to those of the last hidden layer $F_{h_1}$ of the original model, and (ii) a shared unembedding layer $F_u$ from the original model. In Figure \ref{fig:main}c, layers $F_{h_k}$ are illustrated in blue, while $F_u$ is below them. Layer $F_u$ is shared due to the prohibitively large vocabulary sizes of existing LLMs. Given the token context $x_{1:t} =x_1,...,x_t$, each head's input is the hidden representation $z_{1:t}$ from the shared transformer layers $F_s$ of the original model (which excludes $F_{h_1}$). Formally, to output $p_{t+k}$, the $k$-th head is defined as:
\begin{equation} \label{eq:head}
p_{t+k} = \text{softmax}(F_u(F_{h_k}(z_{1:t})))
\end{equation}

Full fine-tuning is used to train layers $F_{h_k}$ for $k>1$, while all other layers are frozen, including the unembedding layer $F_u$. This prevents the performance of layer $F_{h_1}$ from degrading while simultaneously reducing compute costs. The cross-entropy loss for the next $n$ future tokens is:
\begin{equation} \label{eq:aux_head_loss}
\mathcal{L}_n = \sum_{k=2}^{n} - \alpha^{k-2} \log p_{t+k} (y_{t+k})
\end{equation}

where $y_{t+k}$ is the ground truth token at position $t+k$ and $\alpha^{k-1}$ is a geometric decay that lowers the loss of auxiliary heads at later token positions. The higher the future token position, the greater the loss due to its inherent unpredictability. $\alpha^{k-1}$ scales the respective losses to promote more stable training.

In the absence of the original training recipe of most open-source models, we construct an instruction-tuning dataset of 100,000 samples, sourced from the ShareGPT dataset\footnote{Collected from https://sharegpt.com/.} and Tulu 3 SFT mixture \citep{lambert2024t}. It encompasses a broad spectrum of tasks to ensure that the auxiliary heads are task-agnostic; a full breakdown can be found in Table X. Importantly, to match the output distribution of the 1st (original) head, the dataset's ground truth responses are self-distilled from the original head $F_{h_1}$. In other words, only the questions from the dataset are externally sourced.

\subsection{Concept-aware fine-tuning using auxiliary heads}
\label{subsec:method_ft}

After adding the auxiliary heads as trained and defined in Equation \ref{eq:head}, task-specific CAFT can be executed. Generally, only parameters that are part of the original model are fine-tuned. For example, for LoRA fine-tuning, all layers are targeted except for layers $F_{h_k}$ for $k>1$ to reduce memory footprint and except for the unembedding layer $L_u$ to improve training stability. Other fine-tuning methods besides full and LoRA fine-tuning can also be used in theory, but are beyond the scope of this work.

Ultimately, the primary objective is to minimize the 1st head's loss $\mathcal{L}_1$, considering only the 1st head will be used for inference; the losses of all subsequent heads are purely auxiliary. Guided by this motivation, the cross-entropy loss for CAFT is calculated as:
\begin{equation}
\mathcal{L}_n = \sum_{k=1}^{n} - \alpha^{k-1} \beta \gamma \log p_{t+k} (y_{t+k})
\end{equation}

where $\beta$ adjusts the weightage of all auxiliary head losses and $\gamma$ adjusts their weightage over iterations. In practice, we find that models tend to optimize for the auxiliary losses at the expense of the first head's loss $\mathcal{L}_1$ due to the former's relatively higher loss. $\beta=0.01$ ensures that the training continues to prioritize the latter, while a decaying sine schedule for $\gamma$ incentivises models to pay greater attention to the auxiliary losses at the start, but ultimately optimize for $\mathcal{L}_1$.

Importantly, the effectiveness of the multi-token setting is directly correlated with the auxiliary head's adaptation to the given task. Using the method described in Section \ref{subsec:method_aux_heads}, the heads are broadly effective for general conversation, coding, and math tasks. However, for tasks with diverse, unpredictable vocabularies, e.g., story writing, and those with unknown formats, e.g., protein sequences, it is helpful to finetune the auxiliary heads specifically as described in Equation \ref{eq:aux_head_loss} for 1 epoch on the given task's training set. The compute cost is minimal, but vastly improves CAFT's effectiveness.

After training, the auxiliary heads are discarded, such that only the base model remains. Thus, there is no necessary additional computational cost or code modifications for model inference.

\subsection{Practical Implementation}

Concept-aware fine-tuning is straightforward to implement for virtually all language models. In practice, the auxiliary heads of popular models will be trained and open-sourced by various research labs and model providers. Building upon the industry-standard Transformers library \citep{wolf2020transformers}, practitioners need only augment their fine-tuning script with several additional lines of code with our open-source library \brandc{\texttt{caft}}. An example implementation in shown in Figure \ref{fig:use_cases}a.

Some tips for practitioners: first, it is best to monitor both $\mathcal{L}_1$ and $\mathcal{L}_n$ as defined in Equation \ref{eq:aux_head_loss} (without $\beta$ or $\gamma$). We should expect that (i) $\mathcal{L}_n$ decreases over epochs, which shows that the model has optimized for the auxiliary losses, and that (i) $\mathcal{L}_n$ is ultimately lower than that of conventional fine-tuning, which shows that the multi-token objective is beneficial. Second, in practice, we find that when $\mathcal{L}_2 > 4.0$, the auxiliary heads are too unreliable to be useful; in which case, the aforementioned head fine-tuning strategy should be used.

\section{Experiments}

We demonstrate the wide-ranging effectiveness of concept-aware fine-tuning (CAFT) on diverse tasks. Beyond traditional LLM tasks like coding, math, and text generation, we also evaluate its performance on scientific tasks, such as molecular generation and de novo protein design. The Llama3-8B-Instruct model is used as our primary case study, considering its ubiquity in fine-tuning tasks. After training its auxiliary heads, the model is adapted using CAFT with full or LoRA settings on the aforementioned tasks. We find that CAFT methods consistently outperform their next-token counterparts.

\subsection{Training Auxiliary Heads}

\begin{figure}
	\centering
	\includegraphics[width=0.5\textwidth]{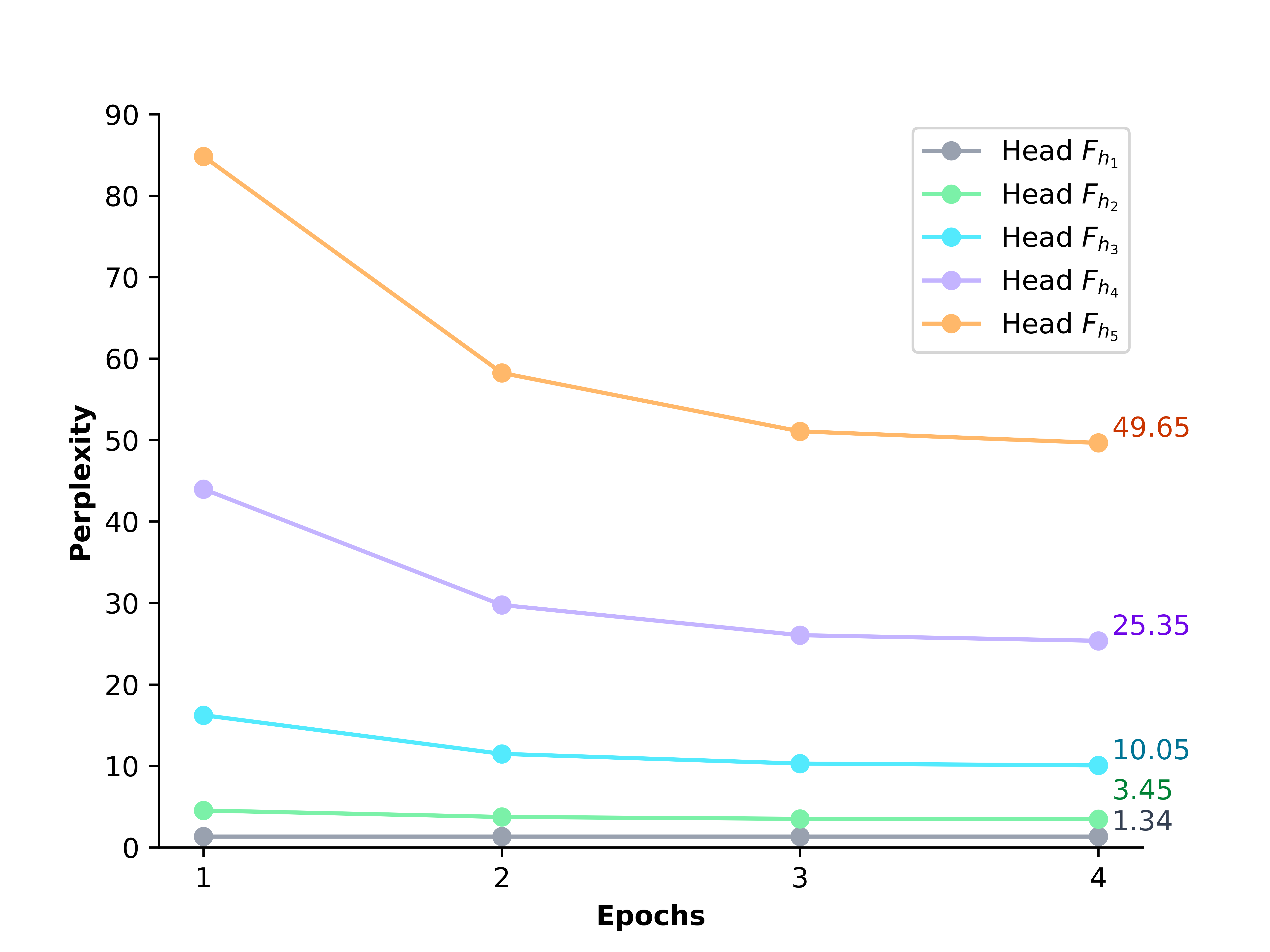}
	\caption{Perplexities of auxiliary heads over four epochs. Note that head $F_{h_1}$ is not adapted and is only displayed for reference.}
	\label{fig:aux_head_ppl}
\end{figure}

We train four auxiliary heads $F_{h_{2:5}}$ for Llama3-8B-Instruct using the method and architecture described in Section \ref{subsec:method_aux_heads}, with a sequence length of 4096 and for up to 4 epochs. $n=5$ is chosen as a balance between the extent of multi-token representation and the additional compute required. Other studies have empirically found that training four additional heads is optimal \citep{gloeckle2024better, cai2024medusa}. As shown in Figure \ref{fig:aux_head_ppl}, the training leads to a significant reduction in model perplexities across all auxiliary heads. It also suggests that four epochs are sufficient to achieve optimal performance. As expected, heads become more inaccurate as the token position increases. Without knowing the predictions of prior heads, subsequent heads will inevitably observe higher losses. Note that the perplexity of the original head $F_{h_1}$ is shown only for reference; it is not trained at this stage.


These task-agnostic auxiliary heads are used to apply CAFT to the downstream tasks below. Note that, in practice, users do not need to train the auxiliary heads and only need to fine-tune on their specific tasks; trained auxiliary heads will be readily available on the internet. 


\subsection{Downstream Tasks}
\label{subsec:tasks_results}

We present the results on five downstream tasks of different domains to demonstrate the superior performance and broad applicability of concept-aware fine-tuning. Example questions from every evaluation dataset can be found in Figure \ref{fig:use_cases}b. The general experimental setup is as follows: using task-specific training datasets, we fine-tune four separate models using full CAFT, LoRA CAFT, and their next-token versions. The latter two serve as baseline models for comparison. They are trained for up to 5 epochs with early stoppage and for sequence lengths ranging from 512 to 2048, depending on the specific task. All results shown are the average of 5 independent evaluation runs and their variance via 95\% confidence intervals (CIs). From Task 3 onwards, the auxiliary heads are pretrained on the training set for 1 epoch before proceeding to the actual fine-tuning.

\subsubsection{Downstream Task 1: Coding}
\label{subsubsec:coding}

Code is an intuitive application of CAFT. Programming languages have vastly different vocabularies and distributions from natural languages. For example, punctuations like brackets and colons convey vastly different semantic meanings. Given the dominance of natural language in pretraining corpora, modern tokenizers do not effectively encode programming-specific texts. Thus, coherent "words" in code, such as \texttt{\_\_name\_\_} in Python, are often deconstructed into two or more unintuitive fragments.

For this experiment, we first construct a Python training dataset that consists of a combined 10,000 examples from CodeAlpaca \citep{codealpaca}, MagiCoder \citep{xu2024wizardlm}, and the Mostly Basic Python Programming (MBPP) training set \citep{austin2021program}. The HumanEval test set \citep{chen2021evaluating} is used to evaluate the models; an example question is shown in Figure \ref{fig:use_cases}b. Models tested under the pass@1 setting, i.e., one response is sampled per question, which is only deemed correct if it passes all test cases.


\begin{table}[ht!]
	\def\arraystretch{1.2}
	\centering
	\caption{Model Performances on HumanEval.}
	\label{table:coding_results}
	\begin{tabular}{llcc}
		\thickhline
        Task & Method & & Accuracy (\%) $\uparrow$ \\
        \thickhline
        \multirow{5}{*}{HumanEval} & Base & & 38.9 \;\sm{$\pm$1.501}\\
        \textbf{} & \multirow[c]{2}{*}{LoRA Fine-tuning} & Next-token & 40.9 \;\sm{$\pm$0.927}\\
        \textbf{} & & \brandc{\textbf{CAFT}} & \textbf{45.1} \;\sm{$\pm$1.930}\\
        \textbf{} & \multirow[c]{2}{*}{Full Fine-tuning} &  Next-token & 40.5 \;\sm{$\pm$2.309}\\
        \textbf{} & & \brandc{\textbf{CAFT}} & \textbf{49.3} \;\sm{$\pm$2.590}\\
        \thickhline
	\end{tabular}
\end{table}

As shown in Table \ref{table:coding_results}, CAFT dramatically improves Python coding performance: LoRA and full CAFT lead to a 4.2\% and 8.8\% improvement in accuracy respectively. The magnitude of these gains is surprising given the relatively small training sample size and short training time. This reinforces our aforementioned hypotheses that the arbitrary parsing of code greatly hinders model learning, suggesting that the current next-token paradigm suppresses the potential of LLMs to complete coding tasks.

\subsubsection{Downstream Task 2: Mathematics}

Similar to code, mathematical expressions are segmented by tokenizers. Most tokenizers have a vocabulary size of 30-120 thousand, which is vastly insufficient to encompass the space of numbers and symbols. A multi-token setting could reassemble these tokens into the relevant expressions.

The math training set consists of 10,000 random samples from MetaMathQA \citep{yu2023metamath}, which are math questions augmented from the \textit{training sets} of GSM8K \citep{cobbe2021training} and MATH \citep{lightman2023let}. Each answer contains the answer steps in the form of chain-of-thought. Models are evaluated on the MATH-500 test set, based on whether their final answer exactly matches the ground truth answer.

\begin{table}[ht!]
	\def\arraystretch{1.2}
	\centering
	\caption{Model Performances on MATH-500.}
	\label{table:math_results}
	\begin{tabular}{llcc}
		\thickhline
        Task & Method & & Accuracy (\%) $\uparrow$ \\
        \thickhline
        \multirow{5}{*}{MATH-500} & Base & & 19.1 \;\sm{$\pm$1.367}\\
        \textbf{} & \multirow[c]{2}{*}{LoRA Fine-tuning} & Next-token & 22.9 \;\sm{$\pm$1.033}\\
        \textbf{} & & \brandc{\textbf{CAFT}} & \textbf{24.6} \;\sm{$\pm$1.666}\\
        \textbf{} & \multirow[c]{2}{*}{Full Fine-tuning} &  Next-token & 23.7 \;\sm{$\pm$0.572}\\
        \textbf{} & & \brandc{\textbf{CAFT}} & \textbf{25.2} \;\sm{$\pm$0.732}\\
        \thickhline
	\end{tabular}
\end{table}

Importantly, this particular experimental setting is highly unfavorable to CAFT. First, the MetaMathQA dataset primarily features natural language reasoning, rather than long, recurring mathematical expressions; the latter, however, benefits the most from our proposed method. Additionally, math problems, unlike the rest of the downstream tasks, are evaluated entirely on their final answers. Thus, CAFT primarily enhances the intermediate chain-of-thought reasoning, thereby only indirectly influencing model accuracy. Nonetheless, as shown in Table \ref{table:math_results}, substantial accuracy gains are observed when CAFT is used.

\subsubsection{Downstream Task 3: Text Generation}

In this task, we explore the use of CAFT for domain-specific text generation, i.e., clinical text summarization. Brief Hospital Course (BHC) summaries are short summaries that describe a patient's hospital stay. By referencing often voluminous clinical notes, BHC summaries are tediously written by clinicians. This significant time burden could be alleviated by leveraging LLMs to create these summaries. 

To evaluate the effectiveness of CAFT for this task, we use MIMIC-IV-BHC \citep{aali2025dataset}, a pre-processed dataset extracted from raw MIMIC-IV notes \citep{johnson2023mimic}. In line with the existing literature on text summarization, we evaluate the models using the ROUGE metrics \citep{lin2003automatic}: ROUGE-N measures the n-gram overlap between the model generation and ground truth summaries, ROUGE-L measures the Longest Common Subsequence (LCS) to captures sentence-level structure and word order, and ROUGE-Lsum calculates the LCS of each sentence for more granularity.

\begin{table}[ht!]
	\def\arraystretch{1.5}
	\centering
	\caption{Model Performances on CNN-DailyMail and WritingPrompts.}
	\label{table:textgen_results}
	\begin{tabular}{lccccc}
		\thickhline
        Method & & ROUGE-1 $\uparrow$ & ROUGE-2 $\uparrow$ & ROUGE-L $\uparrow$ & ROUGE-Lsum $\uparrow$\\
        \thickhline
        \shortstack{Base \\[0pt]} & & \shortstack{\\[0pt] 29.17\\\sm{(0.188)}} & \shortstack{6.64\\\sm{(0.015)}} & \shortstack{15.46\\\sm{(0.048)}} & \shortstack{27.49\\\sm{(0.160)}}\\
        \multirow[c]{2}{*}{LoRA Fine-tuning} & \shortstack{Next-token \\[1pt]} & \shortstack{\\[0pt] 42.31\\\sm{(0.173)}} & \shortstack{20.56\\\sm{(0.136)}} & \shortstack{29.86\\\sm{(0.150)}} & \shortstack{40.44\\\sm{(0.133)}}\\
         & \shortstack{\brandc{\textbf{CAFT}} \\[1pt]} & \shortstack{\\[0pt] \textbf{44.16}\\\sm{(0.277)}} & \shortstack{\textbf{22.30}\\\sm{(0.212)}} & \shortstack{\textbf{31.62}\\\sm{(0.176)}} & \shortstack{\textbf{42.37}\\\sm{(0.288)}}\\
        \multirow[c]{2}{*}{Full Fine-tuning} & \shortstack{Next-token \\[1pt]} & \shortstack{\\[0pt] 44.57\\\sm{(0.219)}} & \shortstack{22.94\\\sm{(0.231)}} & \shortstack{32.17\\\sm{(0.272)}} & \shortstack{42.75\\\sm{(0.218)}}\\
         & \shortstack{\brandc{\textbf{CAFT}} \\[1pt]} & \shortstack{\\[0pt] \textbf{45.93}\\\sm{(0.315)}} & \shortstack{\textbf{24.44}\\\sm{(0.281)}} & \shortstack{\textbf{33.76}\\\sm{(0.306)}} & \shortstack{\textbf{44.04}\\\sm{(0.327)}}\\
        \thickhline
	\end{tabular}
\end{table}

As shown in Table \ref{table:textgen_results}, CAFT methods consistently outperform their next-token counterparts. Unlike tasks like math and coding, text generation in uncontrolled; it exhibits significantly larger linguistic diversity, ranging from natural language text to domain-specific acronyms. Given the relative sparsity of multi-token concepts, there was a concern that the model would fail to learn these representations. However, the clear improvement in performance shows that the auxiliary heads effectively capture ideas spanning multiple tokens, even when they are not as prevalent.

\subsubsection{Downstream Task 4: Molecular Generation}
\label{subsubsec:molgen}

The de novo molecule generation task is defined as follows: Given a set of desired molecular functions and characteristics, generate the corresponding molecular structure. The task of formulating novel molecules has important downstream applications in areas such as drug discovery and materials design, and conventionally relies on the intuition of chemists \citep{meyers2021novo}. While there have been many attempts to use generative models to accelerate the process, they still have significant room for improvement, given the unique grammar of SMILES sequences and the inherent difficulty of associating molecular properties with the given representation. For example, molecular structures often contain functional groups, which are coherent subsequences within the SMILES representation. CAFT presents a unique opportunity to better understand its idiosyncratic syntax and exploit these domain-specific patterns.

For this case study, the train and test sets are drawn from the challenging L+M-24 dataset \citep{edwards2024lpm}. The input is a natural language description of the desired molecular properties, and the output is the associated molecule in the form of a SMILES sequence. The following metrics are used: (i) exact match (EM), representing the proportion of generated SMILES sequence that exactly matches the gold truth, (ii) Levenshtein distance \citep{miller2009levenshtein}, measuring the statistical distance between the model generation and ground truth, (iii) RDK Fingerprint Tanimoto Similarity (FTS) \citep{schneider2015get}, measuring the structural similarity between SMILES strings, and finally (iv) validity, representing the proportion of generations that conform to SMILES grammar and chemical rules.

\begin{table}[ht!]
	\def\arraystretch{1.2}
	\centering
	\caption{Model Performances on L+M-24.}
	\label{table:molgen_results}
	\begin{tabular}{lccccc}
		\thickhline
        Method & & EM (\%) $\uparrow$ & Levenshtein $\downarrow$ & RDK FTS $\uparrow$ & Validity (\%) $\uparrow$\\
        \thickhline
        \shortstack{Base \\[0pt]} & & \shortstack{\\[0pt] 0.00\\\sm{(0.000)}} & \shortstack{139.19\\\sm{(3.621)}} & \shortstack{9.67\\\sm{(0.383)}} & \shortstack{68.70\\\sm{(1.563)}} \\
        \multirow[c]{2}{*}{LoRA Fine-tuning} & \shortstack{Next-token \\[1pt]} & \shortstack{\\[0pt] 0.00\\\sm{(0.000)}} & \shortstack{64.14\\\sm{(3.284)}} & \shortstack{62.42\\\sm{(0.178)}} & \shortstack{91.84\\\sm{(1.019)}}\\
        & \shortstack{\brandc{\textbf{CAFT}} \\[1pt]} & \shortstack{\\[0pt] 0.00\\\sm{(0.000)}} & \shortstack{\textbf{58.69}\\\sm{(3.661)}} & \shortstack{\textbf{63.20}\\\sm{(0.362)}} & \shortstack{\textbf{93.46}\\\sm{(1.158)}}\\
        \multirow[c]{2}{*}{Full Fine-tuning} & \shortstack{Next-token \\[1pt]} & \shortstack{\\[0pt] 0.14\\\sm{(0.068)}} & \shortstack{47.59\\\sm{(2.565)}} & \shortstack{65.34\\\sm{(0.126)}} & \shortstack{92.38\\\sm{(0.778)}}\\
        & \shortstack{\brandc{\textbf{CAFT}} \\[1pt]} & \shortstack{\\[0pt] \textbf{0.54}\\\sm{(0.068)}} & \shortstack{\textbf{41.23}\\\sm{(0.801)}} & \shortstack{\textbf{67.79}\\\sm{(0.172)}} & \shortstack{\textbf{97.14}\\\sm{(0.655)}} \\
        \thickhline
	\end{tabular}
\end{table}

Scientific LLMs typically rely on full fine-tuning, rather than LoRA, due to the complexity of biomedical tasks and lack of exposure to relevant content in their original training corpus. As such, we focus on the results of full CAFT in Table \ref{table:molgen_results}, which offers significant gains over the baseline. This is especially true for exact match accuracy, which improved several-fold, and the percentage of valid SMILES strings, which improved by 4.76\%. These results suggest that CAFT is effective for tasks with usual, out-of-distribution targets.

\subsubsection{Downstream Task 5: De Novo Protein Design}

The goal of de novo protein design is to generate proteins that exhibit the desired functions and characteristics. In conventional protein design, a protein backbone structure at the atomic level is first defined, followed by finding a sequence consistent with that structure. \textit{De novo} protein design, on the other hand, generates the protein from scratch, without relying on existing sequences or other starting points. This method holds immense promise for protein engineering, potentially paving the way for creating proteins with novel architectures and functions, and enabling precise control over the proteins' functions and characteristics \citep{kortemme2024novo}. However, de novo design is deeply challenging given the immense space of potential sequences and unintuitive grammar of protein sequences.

To evaluate CAFT's effectiveness in de novo protein design, we leverage a user requirement-protein sequence pair dataset using features from the UniProt knowledgebase, curated by Mol-Instructions \citep{fang2023mol}. The protein design requirements, expressed in natural language, consist of the protein's general functions, associated metabolic pathways, co-factors, and 17 other commonly targeted features. Due to the significant difficulty of protein sequences, LoRA methods fail to achieve substantial improvements; only fine-tuning methods are reported.

After fine-tuning on 10,000 samples, models are evaluated along two axes: sequence and structural similarity. For the former, we use (i) sequence identity \citep{rost1999twilight}, which measures the proportion of exact matches between the model generated and ground truth sequences, and (ii) sequence alignment, which measures the degree of similarity between sequences using the BLOSUM62 substitution matrix for protein alignments \citep{henikoff1992amino} and the Needleman-Wunsch global pairwise alignment algorithm \citep{needleman1970general}. For structural similarity, the relevant metrics are (i) predicted local distance difference test (pLDDT) \citep{jumper2021highly}, which measures a model's confidence in the protein structure generated from a given sequence, and (ii) TM-score \citep{zhang2004scoring}, which measures the similarity between the model generated and ground truth structures. Note that for the structural metrics, we leverage ColabFold \citep{mirdita2022colabfold}, a fast, popular implementation of AlphaFold2 \citep{jumper2021highly}, to generate the protein structure of the model-generated sequences. pLDDT is important because it measures whether a given sequence can lead to a plausible protein structure. 

\begin{table}[ht!]
	\def\arraystretch{1.2}
	\centering
	\caption{Model Performances on Mol-Instructions De Novo Protein Design.}
	\label{table:proteind_results}
	\begin{tabular}{lccccc}
		\thickhline
        Method & & Identity (\%) $\uparrow$ & Alignment $\uparrow$ & pLDDT $\uparrow$ & TM-score (\%) $\uparrow$\\
        \thickhline
        Base & & \shortstack{\\[0pt] 12.06\\\sm{(0.219)}} & \shortstack{-71.59\\\sm{(3.680)}} & \shortstack{22.28 \\\sm{\phantom{X}}} & \shortstack{6.80 \\\sm{\phantom{X}}}\\
        \multirow[c]{2}{*}{Full Fine-tuning} & \shortstack{Next-token \\[1pt]} & \shortstack{\\[0pt] 20.32\\\sm{(0.180)}} & \shortstack{-16.01\\\sm{(0.208)}} & \shortstack{52.60 \\\sm{\phantom{X}}} & \shortstack{33.07 \\\sm{\phantom{X}}}\\
        & \shortstack{\brandc{\textbf{CAFT}} \\[1pt]} & \shortstack{\\[0pt] \textbf{22.14}\\\sm{(0.180)}} & \shortstack{\textbf{3.18}\\\sm{(1.369)}} & \shortstack{\textbf{54.30} \\\sm{\phantom{X}}} & \shortstack{\textbf{35.12} \\\sm{\phantom{X}}} \\
        \thickhline
	\end{tabular}
\end{table}

As shown in Table \ref{table:proteind_results}, CAFT consistently outperforms the baseline across both axes. In particular, multi-token training leads to significantly higher sequence alignment. Additionally, the proportion of structurally similar protein generations has greatly increased. 25.0\% of generated sequences have a high pLDDT score (typically defined as $>$70.0) for the CAFT model, compared to the 20.0\% of the next-token model; similarly, 20.0\% have a high TM-score (typically defined as $>$50.0) as compared to 15.8\%.

\subsection{CAFT improves performance by learning multi-token concepts.}
\label{subsec:concepts}

We theorize that CAFT's outperformance can be attributed to better conceptual understanding across tokens. An authoritative investigation into the internal mechanisms of CAFT models is, however, unfeasibly difficult. Thus, we support our hypothesis by empirically studying the heterogeneous effects of concepts on CAFT's performance gains. 

\begin{figure}
	\centering
	\includegraphics[width=1.0\textwidth]{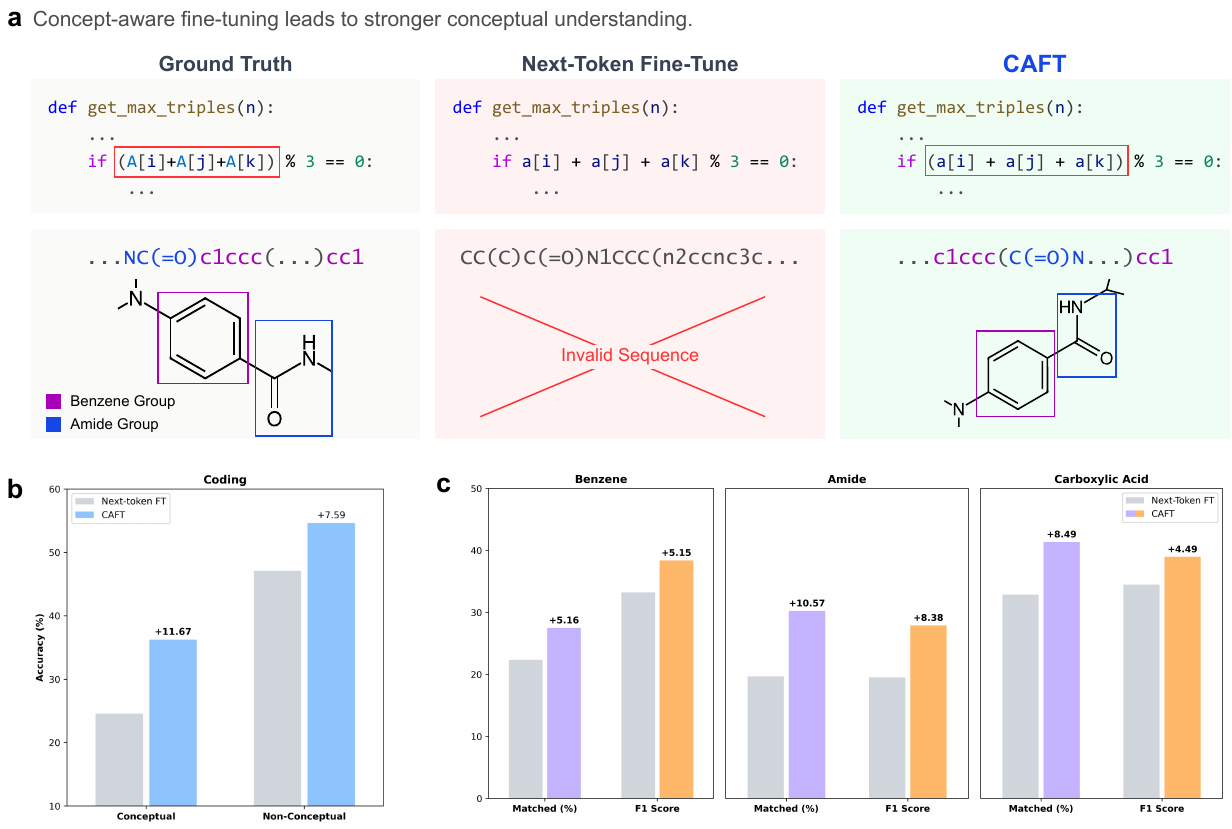}
	\caption{\textbf{(a) Examples of CAFT's concept-informed generation.} A comparison of the ground truth, next-token fine-tuned model generation, and CAFT model generation for two questions from the HumanEval (top) and L+M-24 (bottom) datasets. The red boxes show the relevant proxy concepts, derived using the method described in Section \ref{subsec:concepts}, and how CAFT is able to identify them. \textbf{(b) Coding (HumanEval) Ablation.} The full CAFT model performs disproportionately better on coding questions with a high number of concepts. \textbf{(c) Molecular Generation (L+P-24) Ablation.} The full CAFT model is substantially better at generating the relevant functional groups when required.}
	\label{fig:concept}
\end{figure}

First, we identify \textit{proxies} of concepts in two tasks, coding and molecular generation, using the full CAFT and next-token models from Section \ref{subsubsec:coding} and \ref{subsubsec:molgen}. For Python code, we define concepts as coherent snippets of code that span multiple tokens. In practice, we use a Python parser to extract expressions contained within brackets, quotation marks, etc, and methods separated by periods. For SMILES strings, we define concepts as complex functional groups. This analysis focuses on benzene, amide, and carboxylic acid groups, which are complex yet well-represented in the dataset. Figure \ref{fig:concept}a illustrates two examples from the HumanEval and L+M-24 model evaluations, which show how concepts are extracted from these domains, and how CAFT enables better conceptual generation.

The analysis is conducted as follows: For HumanEval, questions are split into two bins, conceptual or non-conceptual, based on whether their number of concepts is above or below the dataset average. CAFT's performance improvement over next-token fine-tuning across both bins is then measured. For L+M-24, we compare the CAFT and next-token fine-tuned models' ability to correctly generate functional groups. Besides the proportion of matches, to account for false positives, we also report the F1 score.

HumanEval's results, as reported in Figure \ref{fig:concept}(left), show that CAFT's performance gains are more pronounced in highly conceptual questions (+11.67\%) as opposed to less conceptual ones (+7.59\%). L+M-24's results, as shown in Figure \ref{fig:concept}(right), show that CAFT substantially outperforms next-token fine-tuning in learning concepts, i.e., functional groups. The empirical evidence is consistent with our hypotheses that CAFT leads to better multi-token conceptual understanding, but further exploration is required for conclusive prove.












\section{Related Works}

\textbf{Large language model optimization.} LLMs are generally trained to predict the next token autoregressively: Given a sequence $x_{1:t}$, predict the next token $x_{t+1}$. The foundations of this approach were first introduced by the seminal work of \cite{shannon1948mathematical}, and have since grown to form the core of modern LLMs \citep{devlin2019bert, brown2020language}. The mechanism for model optimization is as follows: the model's raw outputs are passed through a softmax activation function to compute its probability distribution for the predicted $x_{t+1}$. It is then compared against the ground truth one-hot probability distribution to compute the cross-entropy loss, which is used to optimize the model weights through backpropagation. Modern LLMs are constructed based on the decoder-only Transformer architecture \citep{vaswani2017attention, radford2018improving}.

\textbf{LLM training pipeline.} There are generally two main phases of LLM training \citep{li2024pre}: pretraining and post-training. During pretraining, models are trained on a massive text corpus in an unsupervised fashion. The goal is to teach language modeling skills and general knowledge. In practice, this phase is responsible for the majority of total compute used. The resulting models, such as DeepSeek V3 \citep{liu2024deepseek} and Llama3-8B-Base \citep{grattafiori2024llama}, serve as "base" models for the next phase: post-training, where they are further trained on supervised datasets to learn specific skills and output formats. This is typically done via supervised fine-tuning and reinforcement learning \citep{ouyang2022training}, where the former consumes the majority of the compute cost in this phase.

\textbf{Downstream Fine-tuning.} Importantly, foundational models are often further trained by industry practitioners and researchers to perform domain-specific tasks, such as math \citep{liu2024acemath}, reasoning \citep{chen2025justlogic}, and even molecular generation \citep{fang2023mol}. However, because fine-tuning all model parameters may be too computationally expensive, parameter-efficient fine-tuning methods such as LoRA \citep{hu2022lora} and QLoRA \citep{dettmers2023qlora} were introduced.

\textbf{Multi-token prediction.} A growing body of literature finds that the next-token training paradigm performs poorly on lookahead tasks \citep{bachmann2024pitfalls} and compositional tasks \citep{dziri2023faith}, and is highly inefficient relative to human children \citep{frank2023bridging}. In response, two existing works introduce multi-token training to the \textit{pretraining} phase \citep{gloeckle2024better, liu2024deepseek}. Prior to our work, multi-token training in the \textit{post-training} phase saw worse performance than the conventional next-token setting \citep{cai2024medusa, gloeckle2024better}. Orthogonal to these works, several speculative decoding methods leverage multi-token prediction to serve as drafts for future tokens \citep{stern2018blockwise, cai2024medusa}. Their primary purpose is to improve inference speed and generally observe minor performance degradation. Our proposed method, CAFT, builds upon both bodies of literature, especially \cite{cai2024medusa} and \cite{gloeckle2024better}, to unlock multi-token prediction's potential for fine-tuning.

\textbf{Concept-based Learning.} Several works have sought to incorporate concepts into neural networks. Concept bottleneck models are trained to predict human-labeled concepts, which are then used for final prediction; these are typically used for computer vision tasks \citep{koh2020concept, kumar2009attribute}. The Large Concept Model \citep{barrault2024large}, defining sentences as concepts, performs next-sentence prediction in an embedding space. Finally, recent works on training LLMs to reason in continuous latent spaces \citep{hao2024training, zhang2025soft} also draw inspiration from conceptual thinking.


\section{Discussion}

\subsection{The Advantages of Concept-Aware Fine-Tuning (CAFT)}

Concept-aware fine-tuning (CAFT) addresses LLMs' fragmented conceptual understanding, caused by the artificial nature of tokenization's text parsing. Specifically, CAFT is the first method to introduce multi-token prediction to model fine-tuning. This was previously thought to be unfeasible due to the dramatic distribution shift and brevity of the post-training phase. CAFT resolves these problems by introducing two main novel techniques: (i) pretraining the auxiliary heads on a general instruction tuning dataset and (ii) dynamically scaling the auxiliary losses. Empirically, CAFT leads to substantial performance gains by improving concept formation.

Our proposed method is designed to be broadly applicable and straightforward to implement. \textbf{(1)} CAFT can be applied to any generation task, including those requiring modalities beyond natural language, such as protein sequences. It is therefore useful for both industry practitioners and researchers. \textbf{(2)} CAFT is cost-efficient. Existing multi-token pretraining methods are prohibitively expensive and can only be adopted by AI hyperscalers, such as Meta and DeepSeek. For the first time, CAFT democratizes the multi-token setting for any company or research lab. \textbf{(3)} CAFT is easy to implement. Using our open-source library \brandc{\texttt{caft}}, only a few additional lines of code must be added to one's existing fine-tuning script.

\subsection{Broader Implications}

\textbf{Rethinking the next-token objective.} LLMs' ability to plan beyond the next token has long been controversial. Learning across tokens is crucial for coherent text generation and accomplishing a broad spectrum of tasks. The broad success of next-token LLMs, starting from GPT-3 \citep{brown2020language}, strongly suggests that existing models can learn coherent concepts across tokens to some extent. Other empirical works confirm this: for one, \cite{lindsey2025biology} find that LLMs plan beyond the next word when writing poems.

However, on the flip side, the multi-token objective has been empirically shown to improve model performance: \cite{gloeckle2024better} and \cite{liu2024deepseek} have done so for the pretraining phase; our work is the first to do so for the post-training phase. If the next-token setting already adequately incentivises lookahead, planning, and multi-token conceptual understanding, these results should not be observed. The unreasonable effectiveness of CAFT suggests that it addresses a significant limitation in the next-token setting, i.e., the weak conceptual understanding.

Thus far, multi-token training has been held back by its restriction to the pretraining phase and its prohibitive computation costs. Given that CAFT directly addresses these factors, we hope the introduction of CAFT will drive the adoption of the multi-token setting during post-training, potentially becoming the \textit{de facto} training objective for model fine-tuning.

\textbf{Tokenization.} As LLMs continue to evolve, token vocabulary sizes have only increased. This is because larger vocabularies improve the representation of unique, domain-specific text fragments and reduce the splitting of words into subwords. For example, from Llama 2 to 3 \citep{grattafiori2024llama}, vocabulary sizes have increased by \textit{four} times, from 32K to 128K. However, this exponentially increases model sizes and training costs; a token size of 128K causes the unembedding layer to balloon to over 500M parameters. Other tokenization methods have, in parallel, sought to capture coherent word phrases \citep{lai2021lattice, liu2025superbpe}.

We posit that CAFT reduces the reliance on specific tokenization strategies. Through multi-token prediction, CAFT allows models to capture coherent phrases even when each independent token is an arbitrary, incoherent subword. Optimal vocabulary sizes may also be lower than prescribed by existing scaling laws \citep{tao2024scaling}. Empirical studies of these ideas can be pursued in future works.

\section{Conclusion}

Tokenization in LLMs segments text into artificial, incoherent fragments. Given the conventional next-token training objective, models fail to adequately learn multi-token concepts, thus constraining their performance. In response, we introduce concept-aware fine-tuning (CAFT), which enables better conceptual understanding by introducing the multi-token setting to the post-training phase. Empirically, CAFT leads to substantially better model performance across diverse tasks, including coding, text summarization, and de novo protein design.

\bibliography{bibliography}
\bibliographystyle{rusnat}

\newpage
\appendix

\section{Auxiliary head training dataset details}

In the absence of the original training recipe of most open-source models, we construct an instruction-tuning dataset of 100,000 samples, sourced from the ShareGPT dataset and the Tulu 3 SFT mixture \citep{lambert2024t}. The latter is a combination of many data sources, which are broken down in Table \ref{table:aux_head_dataset}. This dataset aims to encompass a broad spectrum of tasks and response formats that the base model has already been trained on, such as multi-turn conversations, mathematics, coding, and instruction following. The ShareGPT dataset is leveraged for its extensive multi-turn conversations, while the Tulu 3 SFT mixture is used for its diverse tasks and proven effectiveness in post-training. Note that only the questions from these datasets are used; the "ground truth" responses are distilled from the given original model to prevent large distribution shifts.

\begin{table}[ht!]
	\def\arraystretch{1.2}
	\centering
	\caption{Sources for Auxiliary Head Training Dataset.}
	\label{table:aux_head_dataset}
	\begin{tabular}{llc}
		\thickhline
        \textbf{Dataset} & \textbf{Source} & \textbf{Samples Used}\\
        \thickhline
        ShareGPT & \cite{zheng2023judging} & 42,239  \\
        Tulu 3 Persona MATH & \cite{lambert2024t} & 12,550 \\
        Evol CodeAlpaca & \cite{luo2023wizardcoder} & 8,933\\
        WildChat GPT-4 & \cite{zhao2024wildchat} & 8,339 \\
        FLAN v2 & \cite{longpre2023flan} & 7,425\\
        NuminaMath-TIR & \cite{li2024numinamath} & 5,311\\
        Tulu 3 Persona GSM & \cite{lambert2024t} & 4,226 \\
        OpenMathInstruct-2 & \cite{toshniwal2024openmath2} & 4,175\\
        Tulu 3 Persona Python & \cite{lambert2024t} & 2,939 \\
        Tulu 3 Persona IF & \cite{lambert2024t} & 2,501 \\
        SciRIFF & \cite{wadden2024sciriff} & 823\\
        TableGPT & \cite{li2023table} & 429\\
        \thickhline
        \textbf{Total} & & 100,000 \\
        \thickhline
	\end{tabular}
\end{table}

Importantly, this post-training phase for the auxiliary heads is relatively computationally inexpensive. For reference, the full Tulu 3 SFT mixture contains 939,344 samples. Additionally, given that only the auxiliary heads are adapted during training, training time and memory usage are further reduced.

\section{Understanding the CAFT hyperparameters}

\begin{figure}
	\centering
	\includegraphics[width=0.5\textwidth]{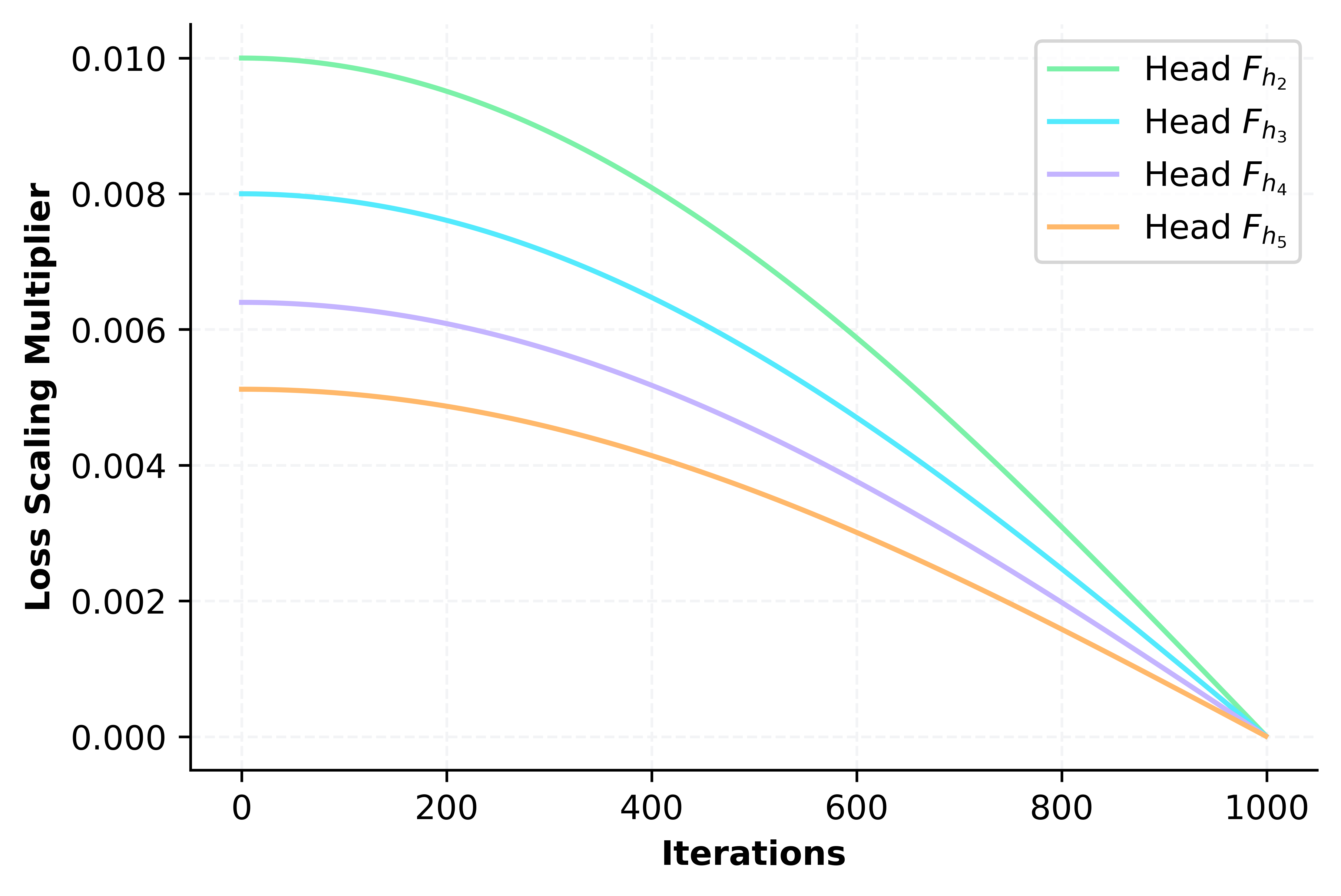}
	\caption{Scaling of training losses of auxiliary heads for $\textrm{iterations}=1000$, $\alpha=0.8$, $\beta=0.01$, and $\gamma$.}
	\label{fig:hyperparams}
\end{figure}

The introduction of the CAFT hyperparameters $\alpha$, $\beta$, and $\gamma$ is key to CAFT's effectiveness. Specifically, they ensure that model's leverage the additional information provided by the auxiliary heads, while ensuring they focus on optimizing $\mathcal{L}_1$. Figure \ref{fig:hyperparams} visualizes the loss scaling of each auxiliary head loss over 1000 iterations, assuming $\alpha=0.8$, $\beta=0.01$, and $\gamma$ is defined by the reflected sine (or RSine) schedule, which is formally:
\begin{equation}
\gamma_t = \sin ((1-\frac{t}{T}) \cdot \frac{\pi}{2})
\end{equation}
where $t$ is the current step and $T$ is the global number of steps. The influence of the auxiliary losses peaks at the start of training and declines over iterations. Additionally, the auxiliary losses are much smaller than the primary $\mathcal{L}_1$, as evidenced by the scaling range of 0.01 to 0. These help ensure that the influence of $L_n$ does not overpower that of $L_1$.

\section{Experiment Settings}

\subsection{Auxiliary Head Training}

We train four auxiliary heads for Llama-3-8B-Instruct using the aforementioned training dataset for 4 epochs, with a maximum sequence length of 4096 and a batch size of 64. For optimization, the 8-bit AdamW optimizer is used with a peak learning rate of 1e-4. To improve training dynamics, 300 warmup steps and a cosine scheduler are implemented.

\subsection{Full and LoRA Concept-Aware Fine-Tuning (CAFT)}


\textbf{Model Training.} All models are trained for 5 epochs with early stoppage and using the AdamW optimizer with a cosine scheduler. 50 warm-up steps are used. Several hyperparameters are task-specific: For sequence length, 512 is used for mathematics, coding, molecular generation, and protein design; 2048 for BHC summarization. Additionally, for BHC summarization, molecular generation, and protein design, the auxiliary heads are specifically trained for 1 epoch before the main fine-tuning.

In general, CAFT is more robust to hyperparameters than conventional fine-tuning. Because CAFT affords more information during training, there is a lower risk of overfitting: Higher learning rates and LoRA rank \& dropout can be used for better performance. For example, full fine-tuning's optimal learning rate is lower than full CAFT's. In practice, we recommend that practitioners start with the same hyperparameters as conventional fine-tuning, and then adjust accordingly if underfitting is observed.

\begin{table}[ht!]
	\def\arraystretch{1.2}
	\centering
	\caption{Hyperparameters for Full and LoRA concept-aware vs. conventional fine-tuning.}
	\label{table:hyperparams}
	\begin{tabular}{lcccc}
		\thickhline
        & \textbf{LoRA FT} & \textbf{LoRA \brandc{CAFT}} & \textbf{Full FT} & \textbf{Full \brandc{CAFT}}\\
        \thickhline
        Epochs & 5 & 5 & 5 & 5 \\
        Peak LR & 1e-5 & 1e-5 & 5e-6 & 1e-5\\
        Batch Size & 32 & 32 & 32 & 32\\
        LoRA Rank & 8 & 8 & - & -\\
        LoRA Alpha & 16 & 16 & - & -\\
        LoRA Dropout & 0.10 & 0.10 & - & - \\
        CAFT $\alpha$ & - & 0.8 & - & 0.8\\
        CAFT $\beta$ & - & 0.01 & - & 0.01\\
        CAFT $\gamma$ & - & RSine & - & RSine\\
        \thickhline
	\end{tabular}
\end{table}

For the CAFT-specific hyperparameters, the objective is to incentivize models to leverage the auxiliary losses and ultimately optimize for $\mathcal{L}_1$. We conducted an extensive hyperparameter search and found the following settings to be robust across all tasks. $\gamma$ is defined using a reflected sine (or RSine) schedule. Empirically, we find that this performs better than a constant or sine schedule. Other hyperparameters are shown in Table \ref{table:hyperparams}. Additionally, when doing a hyperparameter search, we find that $\alpha=0.8$ performs better than 0.7 and 0.9, while $\beta=0.01$ performs better than 0.05 and 0.10. In practice, these values do not need to be further tuned.

\textbf{Datasets.} Every dataset contains 10,000 training samples. The test set sizes are as follows: 164 for HumanEval, 500 for MATH-500, 500 for BHC summarization, 1000 for L+P-24, and 500 for Mol-Instructions protein design. If the original dataset contains more samples than mentioned above, a random subset is extracted. Additionally, a small number of samples are filtered out from the datasets if they do not fit into their respective maximum sequence lengths reported above.

\subsection{Evaluation}

All base, next-token fine-tuned, and CAFT models are provided a zero-shot prompt that gives explicit instructions on the desired output format. Models are asked to generate the relevant code, text, or sequence only, except for MATH-500, where models are asked to ``think step-by-step". Given that early stoppage is used, models are also evaluated on earlier checkpoints, and the top-performing one is reported. 

To control for noise, 5 independent runs are done for every dataset; their average metrics are reported. Additionally, following \cite{grattafiori2024llama}, we report each score's variance via 95\% confidence intervals (CIs). The confidence intervals of pLDDT and TM-score in Table \ref{table:proteind_results} are not reported because the highest scoring output from ColabFold is used instead of the average, as is standard practice in the existing literature.

All the downstream tasks are deterministic, where precision is more important than diversity. Thus, a temperature of 0.1 is used as the default sampling strategy. Only the protein design evaluation deviates from this: all models tend to become repetitive. As such, a temperature of 0.3 and a repetition penalty of 1.1 are used for this task.

\subsection{Experimental Setup of Conceptual Understanding Empirical Study (Section \ref{subsec:concepts})}

To extract multi-token concepts from the model generations, we first conducted a qualitative analysis to identify common ways in which concepts are expressed in the ground truth Python solutions in HumanEval. We found the following types of text fragments to be reasonable proxies: (i) text contained within brackets, e.g. \texttt{(1 + 1)}, (ii) text contained within quotations, e.g. \texttt{"\_\_main\_\_"}, and (iii) objects separated by periods, e.g. \texttt{json.loads}. These fragments are extracted via a Python parser; the number of fragments corresponds to the number of concepts in a question. Questions are then split into two bins, separated by the mean concepts per question of 5. 

As for molecular generation, we define functional groups as chemical concepts within a given molecular structure. Benzene, amide, and carboxylic acid groups are chosen for their relatively long sequence lengths and sufficient frequency in the dataset. The matched (\%) metric is calculated based on whether the model generation contains the same number of a given functional group as the ground truth, given that the ground truth contains at least 1 group. To account for false positives (model generates functional groups for questions that do not contain them), the F1 score is also reported. 









\end{document}